\documentclass[10pt,journal,compsoc]{IEEEtran}
\usepackage{cite}
\usepackage{amsmath,amssymb,amsfonts}
\usepackage{algorithm,algorithmic}
\usepackage{graphicx}
\usepackage{textcomp}
\usepackage{comment}
\usepackage{threeparttable}
\usepackage{hyperref}
\usepackage{threeparttable}
\usepackage{multirow}
\usepackage{booktabs}
\usepackage{subcaption}

\def\BibTeX{{\rm B\kern-.05em{\sc i\kern-.025em b}\kern-.08em
    T\kern-.1667em\lower.7ex\hbox{E}\kern-.125emX}}

\begin{document}
\title{A Multi-Stage Temporal Convolutional Network for Volleyball Jumps Classification Using a Waist-Mounted IMU}
\author{Meng~Shang,
        Camilla~De Bleecker,
        Jos~Vanrenterghem,
		Roel~De Ridder,
		Sabine~Verschueren,
		Carolina~Varon,
		Walter~De Raedt,
        and~Bart~Vanrumste
\thanks{M. Shang, C. Varon, and B. Vanrumste are with KU Leuven, STADIUS, Department of Electrical Engineering, 3000 Leuven, Belgium, e-mail: meng.shang@kuleuven.be.}
\thanks{M. Shang and W. De Raedt are with Imec, Kapeldreef 75, 3001 Leuven, Belgium.}
\thanks{M. Shang and B. Vanrumste are with KU Leuven, e-Media Research lab.}
\thanks{C. De Bleecker and R. De Ridder are with Department of Rehabilitation Sciences, Ghent University, Ghent, Belgium}
\thanks{C. De Bleecker, J. Vanrenterghem, and S. Verschueren are with Department of Rehabilitation Sciences, KU Leuven, Leuven, Belgium.}
}

\maketitle

\begin{abstract}
Monitoring the number of jumps for volleyball players during training or a match can be crucial to prevent injuries, yet the measurement requires considerable workload and cost using traditional methods such as video analysis. Also, existing methods do not provide accurate differentiation between different types of jumps. In this study, an unobtrusive system with a single inertial measurement unit (IMU) on the waist was proposed to recognize the types of volleyball jumps. A Multi-Layer Temporal Convolutional Network (MS-TCN) was applied for sample-wise classification. The model was evaluated on ten volleyball players and twenty-six volleyball players, during a lab session with a fixed protocol of jumping and landing tasks, and during four volleyball training sessions, respectively. The MS-TCN model achieved better performance than a state-of-the-art deep learning model but with lower computational cost. In the lab sessions, most jump counts showed small differences between the predicted jumps and video-annotated jumps, with an overall count showing a Limit of Agreement (LoA) of 0.1±3.40 (r=0.884). For comparison, the proposed algorithm showed slightly worse results than VERT (a commercial jumping assessment device) with a LoA of 0.1±2.08 (r=0.955) but the differences were still within a comparable range. In the training sessions, the recognition of three types of jumps exhibited a mean difference from observation of less than 10 jumps: \textit{block}, \textit{smash}, and \textit{overhead serve}. These results showed the potential of using a single IMU to recognize the types of volleyball jumps. The sample-wise architecture provided high resolution of recognition and the MS-TCN required fewer parameters to train compared with state-of-the-art models.
\end{abstract}


\maketitle

\section{Introduction}
\label{sec:introduction}
\IEEEPARstart{J}{umping} is a crucial behavior during volleyball playing. Players must execute various types of jumps (such as block, smash, and overhead serve) during training sessions and matches to execute tactics and strategies effectively \cite{costa_determinants_2011}. However, frequent jumping can impose significant loads on the joints and increase the risk of developing patellar tendinopathy \cite{van_der_worp_risk_2012}. To minimize these injuries, it is essential to monitor and regulate the frequency of jumps by volleyball players. Furthermore, coaches need to record and analyze players' movements to enhance training programs. Therefore, there is an increasing demand to identify the occurrence and types of jumps during volleyball playing.\par

Video recording is frequently used by coaches to record and monitor volleyball players \cite{chen_ball_2012}. Despite the high reliability, this method requires significant workload and resources for setup and analysis. To develop more efficient systems, research teams and companies are turning to wearable sensors that can track jumping-related activities of athletes. Although popular wearable devices like VERT (VERT, Mayfonk Athletic, FL, USA) and ClearSky T6 (Catapult Sports, Melbourne, Australia) \cite{brooks_quantifying_2021, benson_validation_2020} have been validated in various scenarios, they do not provide recognition of the types of volleyball jumps. Furthermore, the algorithms and model parameters of commercial devices are not publicly available, limiting researchers' ability to explore their full potential.\par

Inertial Measurement Units (IMUs) are widely applied in wearable systems for jumping assessment. Previous studies have shown that accelerometers can be utilized for jump detection, with some studies relying on heuristic methods. For example, a study quantified jump frequency by identifying acceleration peaks \cite{jarning_application_2015}. However, it claimed that the results remained poor and effective methods still needed to be explored and validated. Besides, they do not allow for the recognition of different types of volleyball jumps. \par

The use of machine learning models has significantly improved the accuracy and robustness of activity monitoring. However, while these models have been successful in many activity recognition tasks \cite{demrozi_human_2020}, their application in volleyball jumping has reported overall accuracy lower than 80\% \cite{salim_towards_2020}. Also, as a popular field in machine learning, deep learning models such as Convolutional Neural Networks (CNN) \cite{lee_human_2017} and Recurrent Neural Networks (RNN) \cite{zhao_deep_2018} have not yet been explored in the context of volleyball jumps. \par

This study aims to recognize the occurrence and type of each jump during volleyball playing by a single IMU. Data were collected in the laboratory, and during real training sessions of volleyball players, each wearing an IMU on their waist. Using the collected signals, a sample-wise deep learning model was developed to recognize the different types of jumps. \par

This paper is organized as follows: Section~\ref{sec:related} reviews the state-of-the-art methods for activity recognition. Section~\ref{sec:method} describes the proposed system for recognizing volleyball jumps. Section~\ref{sec:results} presents the recognition performance of the system, and section~\ref{sec:discuss} discusses the obtained results. Finally, section~\ref{sec:conclusion} concludes this study and outlines future work.

\section{Related works}
\label{sec:related}

\subsection{Sample-wise classification systems}
The sliding window technique is commonly applied for Human Activity Recognition (HAR) \cite{demrozi_human_2020}. This technique divides the signals into fixed-length slices, each given a label for machine learning tasks. The window size is determined by the specific activities being recognized and can range from small (e.g., 0.5s \cite{salim_towards_2020} and 1s \cite{hammerla2016deep}) to large (e.g., 30s \cite{bashir2016effect}) or even self-adjustable \cite{qi2019human}. Since each window of the time series is classified as a separate class, this technique is called window-wise classification. However, there are two main drawbacks to this approach. Firstly, a trade-off arises when choosing the window size. A large window contains too much irrelevant information for the target activity, while a small window would capture limited information, resulting in reduced performance. For example, in one study on volleyball jumps, a 0.5-second sliding window was used and the results showed overall accuracy under 80\% \cite{salim_towards_2020}. Secondly, the temporal resolution of recognition is limited using the sliding window technique. Although increasing overlap between windows could increase the resolution, each window is independent and the recognition is not temporally consistent. A commonly used overlap rate is 50\% \cite{wang2019survey}. \par

Sample-wise classification refers to a technique in which both the input and output are a sequence of samples. This technique is popular in video-based recognition \cite{farha2019ms}. Compared with the window-wise technique, it gives each sample a label individually. Recently, some studies have reported progress in sensor-based sample-wise classification \cite{wang2022drinking, filtjens2022skeleton}. Compared with the window-wise technique, each sample from the sensors can generate a corresponding prediction. Therefore, it is more sensitive to the transition of the signals and the resolution for recognition is one sample.

\subsection{Temporal convolutional networks}
Traditionally, HAR systems relied on extracting hand-crafted features from each segmented time series \cite{demrozi_human_2020}. These features were based on prior knowledge about the sensors and activities. After feature extraction, conventional machine learning models such as Support Vector Machines and Random Forests were applied for classification \cite{wang2019survey}.\par

Deep learning models offer an end-to-end classification architecture where the signals could be directly fed into the models for classification without the need for hand-crafted features. Convolutional neural networks (CNNs) are widely applied in HAR systems due to their ability to extract features from adjacent samples for neighboring information \cite{lee_human_2017}. Recurrent neural networks (RNNs), on the other hand, were specifically developed for time series classification and regression, retaining information from a longer range \cite{zhao_deep_2018}. Considering the advantages and limitations of both architectures, researchers successfully combined them to improve classification performance. The combination of CNNs and Long-Short Term Memory (LSTM) resulted in CNN-LSTM models that outperformed previous deep learning models \cite{mekruksavanich_smartwatch-based_2020, mutegeki_cnn-lstm_2020}. However, CNN-LSTM contains more parameters and requires increased computational resources \cite{9115647}.\par

In recent years, Temporal Convolutional Networks (TCNs) have emerged as a promising solution to reduce model complexity while maintaining high performance in time series prediction tasks \cite{farha2019ms}. TCNs are based on the traditional CNN architecture but include residual blocks and dilated convolutions, allowing them to learn from longer time series without increasing the model's complexity. Residual blocks also enable deeper structures without gradient vanishing or exploding. Multi-Stage TCN (MS-TCN) was developed to refine the predicted time series by stacking the networks. MS-TCN has been successfully applied in various fields, including video-based action recognition \cite{farha2019ms, czempiel2020tecno}, gesture recognition \cite{wang2022drinking}, and Surgical phase recognition \cite{filtjens2022skeleton}. However, no previous study has investigated the use of MS-TCN in jump detection. \par

The contributions of this study are:

\begin{itemize}
	\item This is the first study that proves the possibility of detecting some types of volleyball jumps using a single waist-mounted IMU. This unobtrusive system was validated in real-life training sessions of professional volleyball players.
	\item This is the first study to apply MS-TCN in sports-related scenarios. The study proves the advantages of MS-TCN over the other state-of-the-art models by reducing complexity without sacrificing recognition performance.
	\item This study provides new insights into sample-wise jumps recognition that outperforms sliding-window techniques, especially for short-time activities recognition.
	
\end{itemize}

\section{Methodology}
\label{sec:method}

This section introduces the details of the sample-wise MS-TCN model, including the problem statement, implementation, and evaluation.

\subsection{Data collection}

This study was approved by the ethical committee of the University Hospital in Ghent, Belgium (number of approval: BC-07679). Written informed consent was obtained from all participants prior to study participation.\par

The dataset was collected in five different sessions. The participants were either elite or non-elite volleyball players. For the first session, the participants performed different jumps in the lab without a ball. For the other four sessions, the participants (that were different from those in the lab session) were observed during club-based volleyball sessions. Before the sessions, the participants were asked to wear an IMU (Actigraph Link GT9X) at the back at the levels of crista iliaca (L4-L5). In the lab session, as a comparison, the participants also wear a VERT device on the waist for jump counts. During the sessions, they were recorded by a camera (Sony Handycam HDR-PJ410). The data of camera and IMU were synchronized using the global timestamps. In the lab session, the subjects were recorded one by one, while in each training session, the subjects were recorded together. The information of the five sessions is shown in Table~\ref{tab:info}\par

\begin{table*}[!t]
\caption{Information about the lab session and training sessions}
\center
\label{tab:info}
\begin{tabular}{|l|l|l|l|l|l|l|l|}
\hline
\textbf{Session} & \textbf{\begin{tabular}[c]{@{}l@{}}number of \\ subjects\end{tabular}} & \textbf{age} & \textbf{gender} & \textbf{height (cm)} & \textbf{weight (kg)} & \textbf{\begin{tabular}[c]{@{}l@{}}duration of \\ session (min)\end{tabular}} & \textbf{\begin{tabular}[c]{@{}l@{}}number of \\ jumps\end{tabular}} \\ \hline
Lab              & 10                                                                     & 21.30$\pm$4.22 & 5M, 5F          & 184.53$\pm$11.60       & 73.42$\pm$12.29        & \begin{tabular}[c]{@{}l@{}}107.13\\  (in total)\end{tabular}                  & 335                                                                 \\ \hline
Training 1       & 5                                                                      & 29.80$\pm$3.87 & 5M              & 188.18$\pm$6.08        & 85.34$\pm$575          & 120.59                                                                        & \multirow{4}{*}{4423}                                               \\ \cline{1-7}
Training 2       & 9                                                                      & 21.33$\pm$2.43 & 9M              & 178.26$\pm$557         & 72.88$\pm$10.12        & 120.70                                                                        &                                                                     \\ \cline{1-7}
Training 3       & 7                                                                      & \            & 7M              & \                    & \                    & 101.72                                                                        &                                                                     \\ \cline{1-7}
Training 4       & 5                                                                      & 27.75$\pm$6.02 & 4M              & 191.33$\pm$5.56        & 84.60$\pm$6.21         & 117.09                                                                        &                                                                     \\ \hline
\end{tabular}
\end{table*}

\begin{figure}[!t]
     \centering
     \begin{subfigure}[t]{0.13\textwidth}
         \centerline{\includegraphics[width=\textwidth]{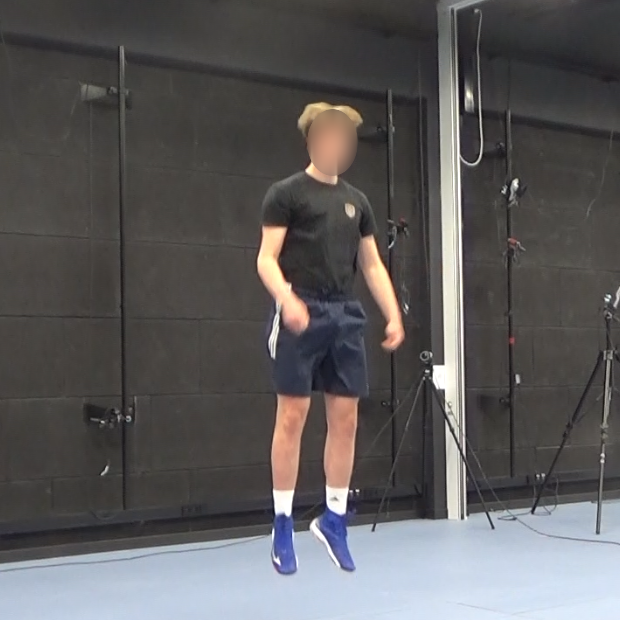}}
         \caption{CMJ}
     \end{subfigure}
     \begin{subfigure}[t]{0.13\textwidth}
         \centerline{\includegraphics[width=\textwidth]{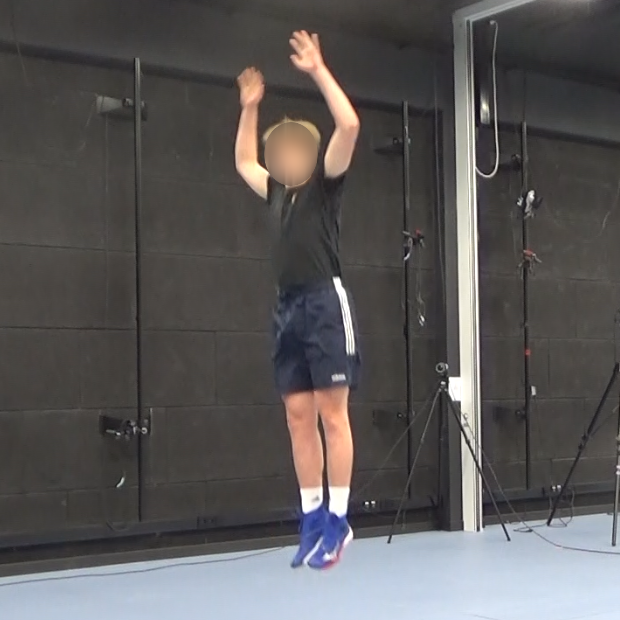}}
         \caption{block}
     \end{subfigure}
     \begin{subfigure}[t]{0.13\textwidth}
         \centerline{\includegraphics[width=\textwidth]{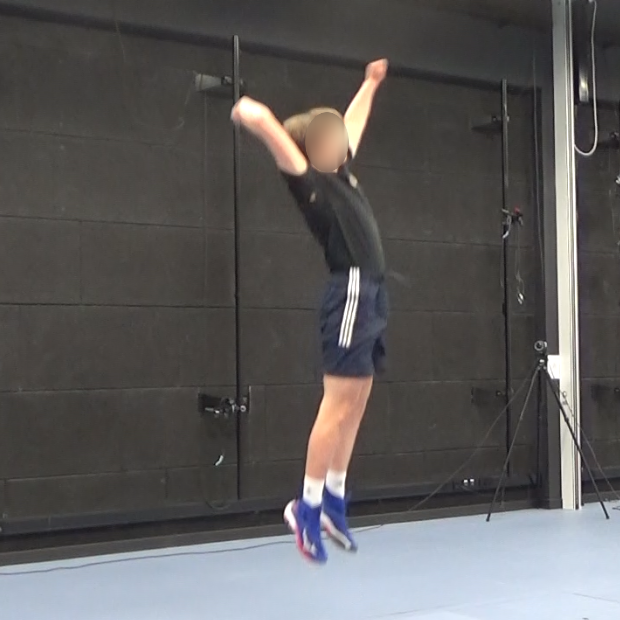}}
         \caption{smash}
     \end{subfigure}
     \begin{subfigure}[t]{0.13\textwidth}
         \centerline{\includegraphics[width=\textwidth]{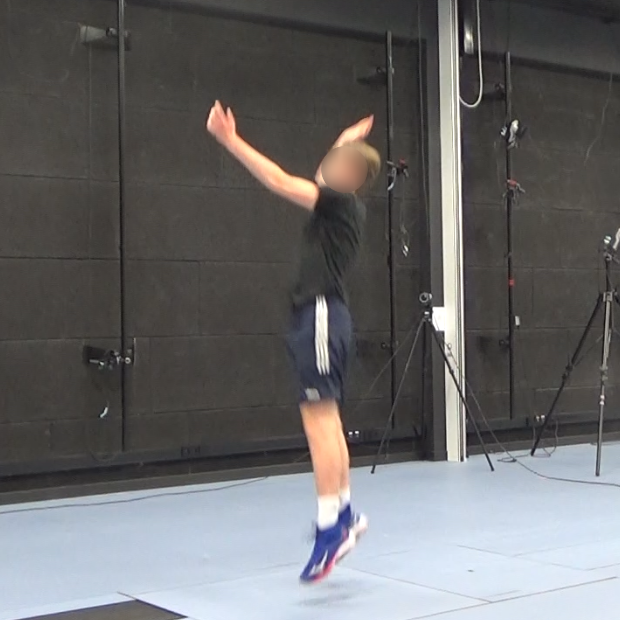}}
         \caption{OS}
     \end{subfigure}
     \begin{subfigure}[t]{0.13\textwidth}
         \centerline{\includegraphics[width=\textwidth]{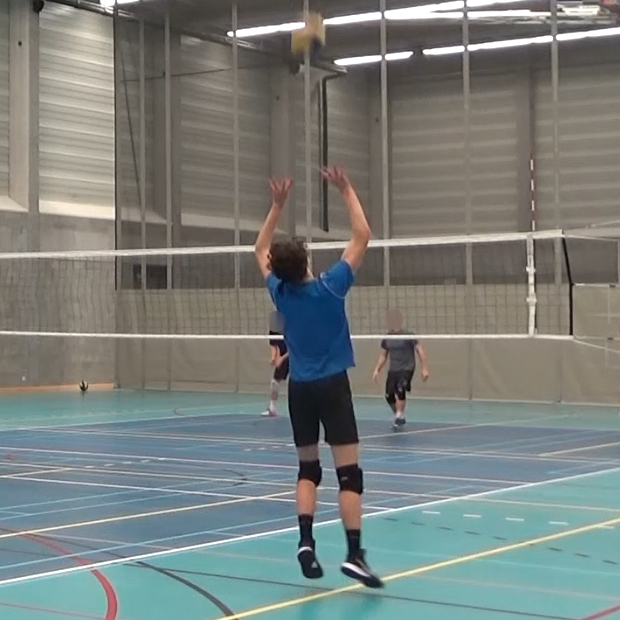}}
         \caption{OP/S}
     \end{subfigure}
     \begin{subfigure}[t]{0.13\textwidth}
         \centerline{\includegraphics[width=\textwidth]{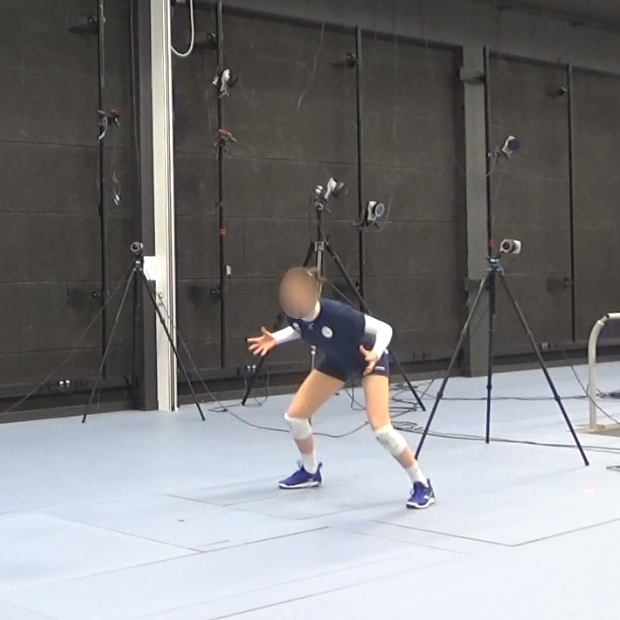}}
         \caption{squat}
     \end{subfigure}
     \begin{subfigure}[t]{0.13\textwidth}
         \centerline{\includegraphics[width=\textwidth]{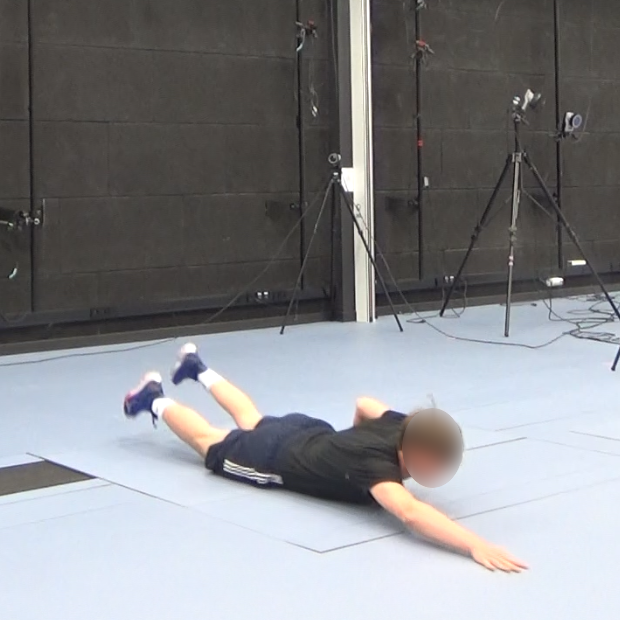}}
         \caption{dive}
     \end{subfigure}
     \begin{subfigure}[t]{0.13\textwidth}
         \centerline{\includegraphics[width=\textwidth]{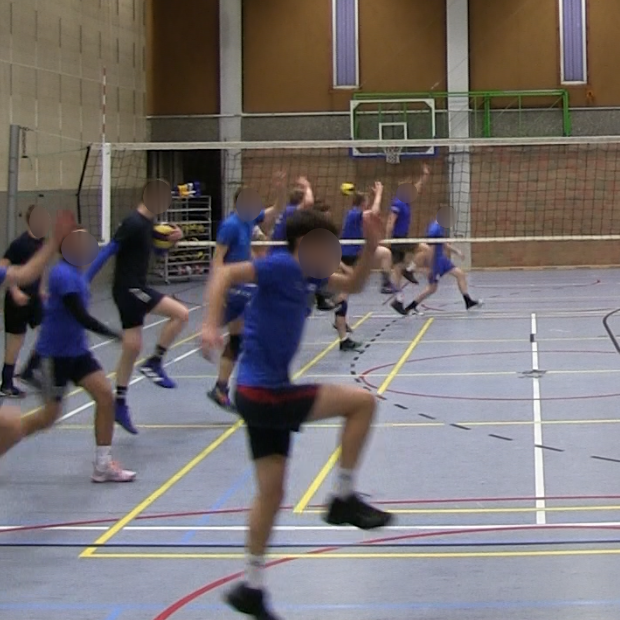}}
         \caption{hop}
     \end{subfigure}
        \caption{The activities performed by the players in the dataset}
        \label{fig:activities}
\end{figure}

During the sessions, they performed different types of volleyball jumps. In session 1 (lab), the participants were instructed to perform the following activities: \textit{Counter Movement Jump (CMJ)}, \textit{block}, \textit{smash}, \textit{overhead serve (OS)}, \textit{squat} and \textit{dive}. For the other sessions, the participants joined their daily training sessions instructed by the coach. The following activities were annotated: \textit{block}, \textit{smash}, \textit{OS}, \textit{overhead pass/set (OP/S)}, \textit{other hops}, \textit{squat} and \textit{dive}. The examples of the performed activities are shown in Fig.~\ref{fig:activities}. \par

\subsection{Problem statement}

In HAR systems, the sensors record the signals during the activities. The recorded signal was a time series of samples $A = \{a_1, a_2, a_3,... a_T\}$. $A$ has a dimension of $T \times N$ where T is the number of samples and N is the number of sensor channels. Using sliding windows, the time series was segmented into some stacks of segments $W = \{w_1, w_2, w_3,... w_P\}$. $W$ has a dimension of $P \times t \times N$ where P is the number of segments and t is the length of the sliding window. Window-wise systems recognize each segment as a label, yielding the output $O_{win} = o_{win1}, o_{win2}, o_{win3},... o_{winP}$. The resolution is therefore $t*(1-r)$, where r is the overlap rate.\par

\begin{figure}[!t]
\centerline{\includegraphics[width=\columnwidth]{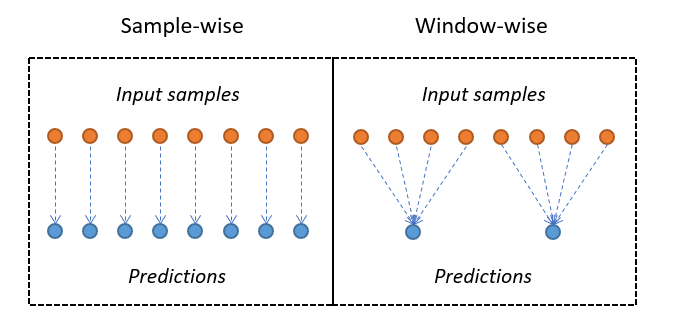}}
\caption{The illustration of window-wise and sample-wise classification}
\label{fig:sample window}
\end{figure}

In this study, a sample-wise system was developed. Sample-wise classification systems predict an outcome for each sample in A. The output is $O_{sam} = o_{sam1}, o_{sam2}, o_{sam3},... o_{samT}$. The resolution is therefore always one sample (full resolution). Considering the short flight time of all volleyball jumps, a high resolution (e.g. less than one second \cite{salim_towards_2020}) is preferred to make sure that each short jump can be captured. Fig.~\ref{fig:sample window} illustrates the concept of window-wise and sample-wise systems.

\subsection{Single-stage TCN}

A single-stage TCN (SS-TCN) is a stack of multiple residual dilated convolution layers \cite{farha2019ms}. Using a dilated layer, the model could learn from larger intervals in time without increasing computational cost. Each dilated convolutional layer was followed by a RELU operation. As shown in Fig.~\ref{fig:MSTCN} (the right part), at the output, a residual connection was applied to avoid the vanishing or exploding gradients problems.\par

\begin{figure}[!t]
\centerline{\includegraphics[width=\columnwidth]{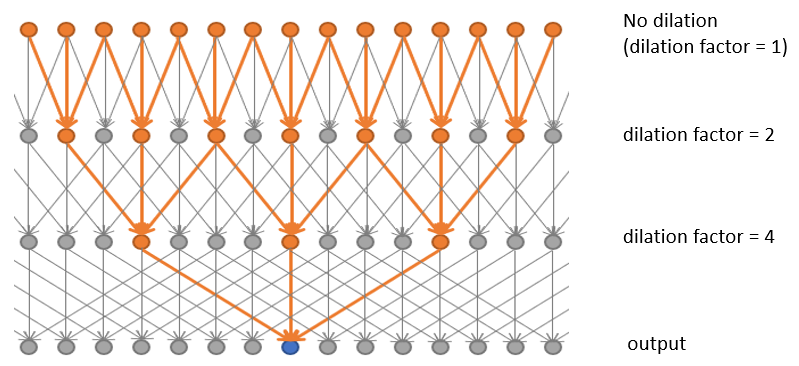}}
\caption{The overview of SS-TCN with casual dilated layers}
\label{fig:SSTCN}
\end{figure}

Each single residual dilated convolution layer was stacked to extend the receptive field. From the first layer (dilation factor = 1), the dilation factors are doubled for each next layer: 2, 4, 8, 16... For easier computation, each dilated layer has a filter of $3\times D$, where D is the number of filters. A single-stage TCN with non-casual dilated convolutional layers is illustrated in Fig.~\ref{fig:SSTCN}. By increasing the dilation factors, the receptive field can be computed as:

\begin{equation}
\text{receptive field} = 2^{L+1}-1,
\label{RF}
\end{equation}

where L is the number of layers in the single stage.\par

In this study, since volleyball jumps were short-time activities, the number of layers was selected as 12 \cite{farha2019ms}. At the output of the last layer, a softmax function was applied to compute the probability of each sample.

\subsection{Multi-stage TCN}

\begin{figure}[!t]
\centerline{\includegraphics[width=0.8\columnwidth]{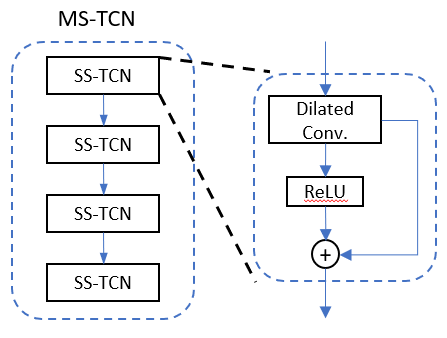}}
\caption{The architecture of MS-TCN and SS-TCN. The left part shows the MS-TCN stacked by SS-TCN. The right part shows the operations in SS-TCN.}
\label{fig:MSTCN}
\end{figure}

A challenge of sample-wise classification is the over-segmentation error. Over-segmentation error occurs when a segment of activity is classified as too many smaller segments of activities. MS-TCN was developed to reduce over-segmentation errors. In MS-TCN, each SS-TCN was stacked, as shown in Fig.~\ref{fig:MSTCN}. From the second stage, each stage takes the output (probability) from the last stage. All stages but the first one served as refinement stages, where both the input and the output were the predicted probability of each sample. Finally, the output of the last stage was used for classification.

\subsection{CNN-LSTM}

\begin{figure}[!t]
\centerline{\includegraphics[width=0.7\columnwidth]{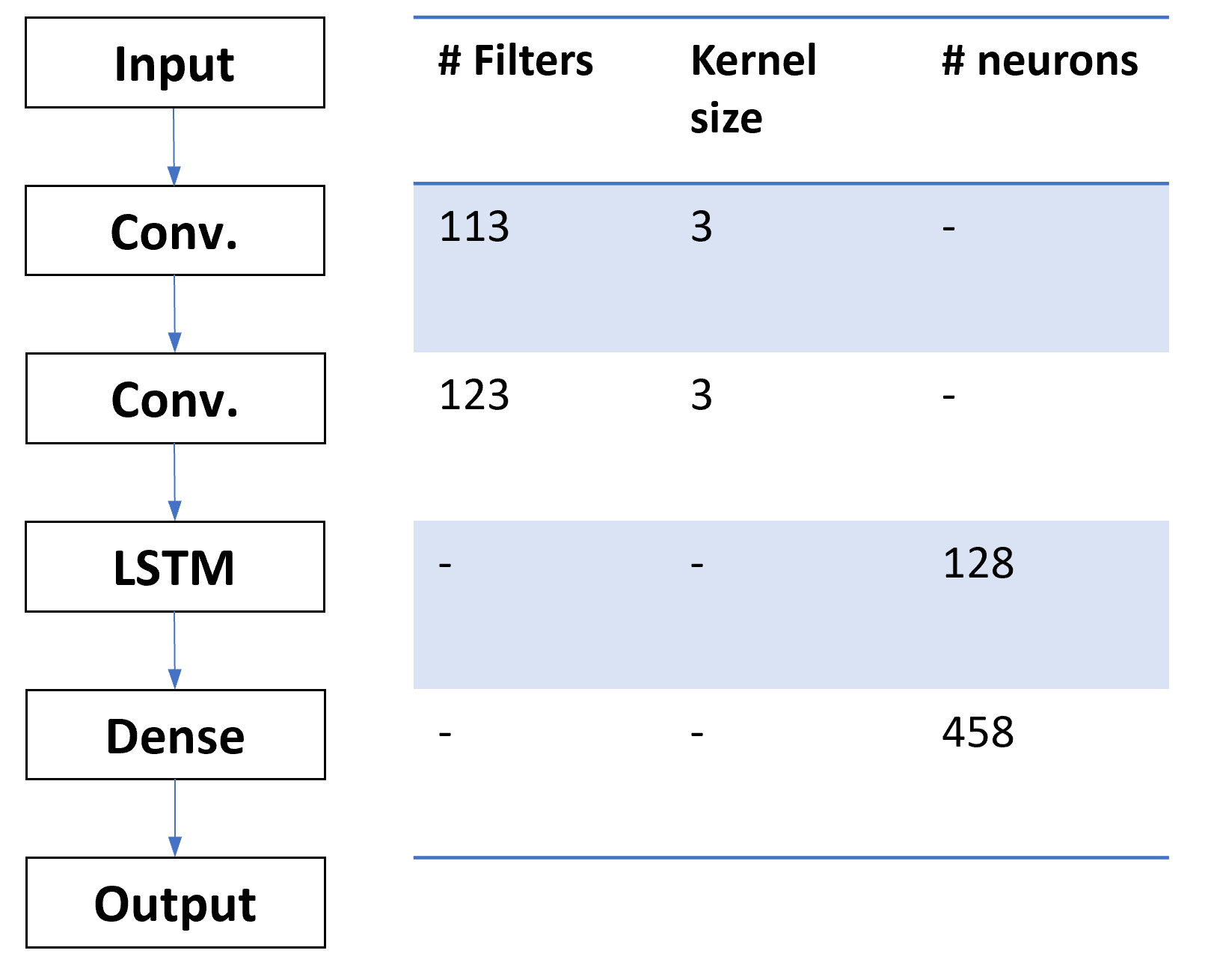}}
\caption{The architecture and hyperparameters of CNN-LSTM}
\label{fig:CNNLSM}
\end{figure}

Since CNN-LSTM is one of the state-of-the-art models for HAR, it was applied as a baseline model. The architecture is illustrated in Fig.~\ref{fig:CNNLSM}. The hyperparameters were set according to a previous study \cite{mekruksavanich2020smartwatch}. Two convolutional layers were first applied to extract features from neighboring samples. To learn from a time series, the stride of the filters was all set to one so that the output of each layer had the same length as the input sequence. Each convolutional layer was followed a dropout layer to avoid over-fitting problems. \par

Afterward, an LSTM layer was applied to the features extracted by the convolutional features. Therefore, output contained long-time temporal information. The LSTM layers also returned a sequence with the same length as the input sequence. A following fully-connected layer classified the probability. \par

\subsection{loss function}

As a classification problem, for every single stage $s$, the Cross Entropy Loss function (CE) was applied for the prediction of each sample, defined as

\begin{equation}
L_{CE}^s=\frac{1}{TC}\sum_{t,c}-y_{t,c}\log{\hat{y}_{t,c}^s},
\label{LCE}
\end{equation}

where $y_{t,c}^s$ denotes the true sample class at time $t$ belonging to class $c$, while $\hat{y}_{t,c}^s$ denotes the predicted probability from stage $s$. $T$ is the number of samples of the sequence and $C$ is the number of classes (number of activities). \par

To further reduce the over-segmentation error, another Truncated Mean Squared Error (TMSE) was added \cite{farha2019ms}. For each stage $s$, the function $L_{TMSE}^s$ is defined as

\begin{equation}
L_{TMSE}^s=\frac{1}{TC}\sum_{t,c}(\Tilde{\Delta}_{t,c}^s)^2,
\label{LTMSE}
\end{equation}

where $\Tilde{\Delta}_{t,c}^s$ is the truncated difference between the loss values of two adjacent samples

\begin{equation}
\Tilde{\Delta}_{t,c}^s=
    \begin{cases}
      \Delta_{t,c}^s, & \text{if $\Delta_{t,c}^s\leq\tau$}  \\
      \tau, & \text{otherwise}
    \end{cases},     
\label{deltat}
\end{equation}

\begin{equation}
\Delta_{t,c}^s = |log{\hat{y}_{t,c}^s}-log{\hat{y}_{t-1,c}^s}|,
\label{delta}
\end{equation}

The TMSE function gave truncated sample-wise difference. By converging the function, the models were less likely to over-segment the time series. The TMSE function and CE function was combined for each single stage

\begin{equation}
L^s = \lambda*L_{TMSE}^s + L_{CE}^s.
\label{loss_s}
\end{equation}

For MS-TCN, the losses of all stages were summed up

\begin{equation}
L = \sum_{s} L^s.
\label{loss_all}
\end{equation}

In this combined loss function, the two hyperparameters were $\lambda$ and $\tau$. In this study, $\lambda$ was set to 0.15 and $\tau$ was set to 4, as optimized in \cite{farha2019ms}.

\subsection{Model selection, training and testing}
\label{subsec:hyper}

\begin{figure}[!t]
\centerline{\includegraphics[width=1\columnwidth]{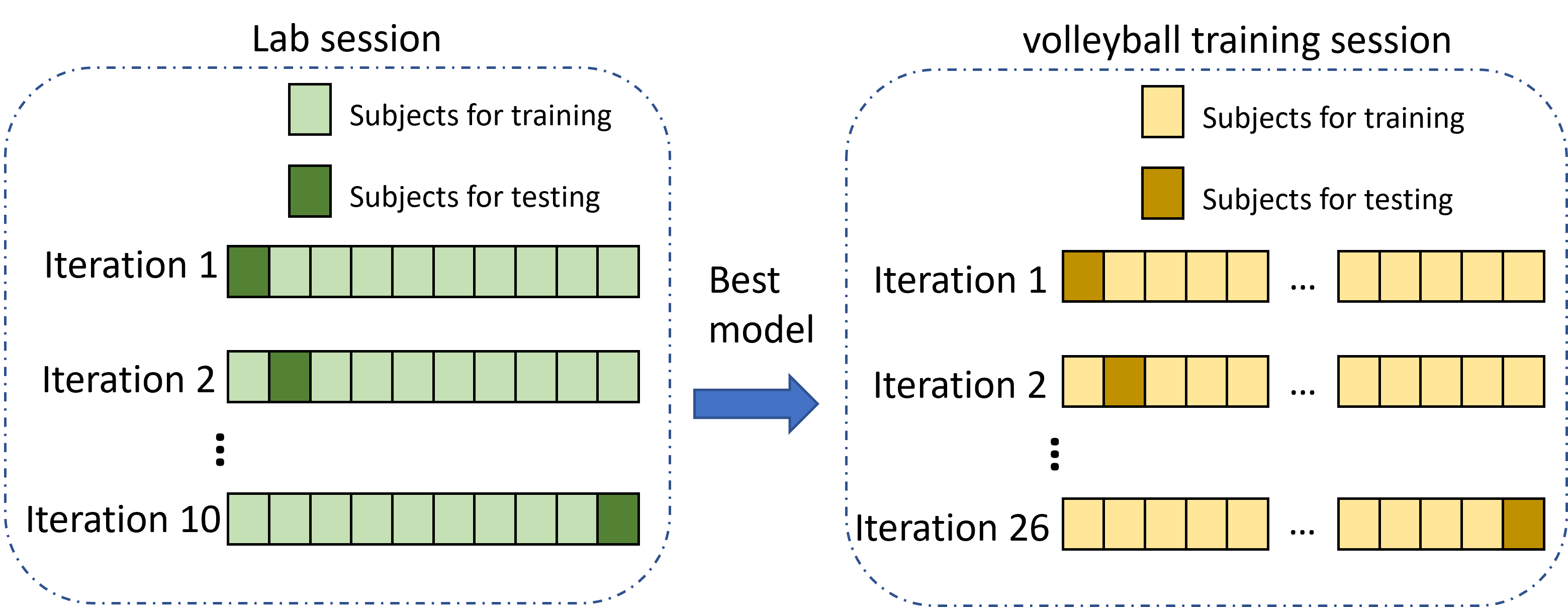}}
\caption{Data splits: the training set and test set in the lab session (for model comparison) and in the volleyball training sessions}
\label{fig:LOSO}
\end{figure}

In this study, most hyperparameters of MS-TCN were selected according to the previous study (number of layers in each stage = 12, $\lambda$ = 0.15, and $\tau$ = 4) \cite{farha2019ms}. The model selection step was performed on the ten subjects from the lab, where CNN-LSTM and MS-TCN (with different number of stages) were compared, as shown in Fig.~\ref{fig:LOSO}.

With the best model selected, LOSO was then applied to split the training set and test set. For each iteration, one subject from the four training sessions was involved in the test set while the other subjects were in the training set. The process is shown in Fig.~\ref{fig:LOSO}.\par

Due to the large size of the dataset, batch training could speed up the training stage and stabilize the optimization process. In the training stage, the time series were cut into slices, with each slice of 40 seconds (the slices should be longer than the receptive field in SS-TCN and MS-TCN). This yields all of the input signals with the same shape for batch training. The batch size was set to 16. In the testing stage, the whole time series was fed into the model without cutting into slices.

\subsection{Evaluation}

The manual annotation was done on the video recordings made at 20 fps, While the IMU had a sampling frequency of 100 Hz. Thus, sample-wise evaluation of the recognition results was not expected to yield high accuracy, especially when transitioning from one activity to another. Therefore, the number of jumping counts was calculated to evaluate the models. By calculating the difference between the true number and predicted number of each type of activity for each subject, Limits of Agreements (LOA) were obtained

\begin{equation}
LOA = M_{pd} + 2*SD_{pd},
\end{equation}

where $M_{pd}$ and $SD_{pd}$ are the mean and standard deviation of the paired difference between the ground truth jump count and predicted jump count

\begin{equation}
M_{pd} = \frac{1}{U}\sum_{u}(nt_{u}-np_{u}),
\end{equation}
\begin{equation}
SD_{pd} = \sqrt{\frac{1}{U-1}\sum_{u}(M_{pd}-np_{u})^{2}},
\end{equation}

where $np_{u}$ and $nt_{u}$ are the predicted jump counts and true jump counts, respectively, for a certain subject u, with U denoting the total number of subjects. LOA, $M_{pd}$, and $SD_{pd}$ could be calculated for each jump type and for all jumps together.

\section{Results}
\label{sec:results}

\subsection{Model selection}

Table~\ref{tab:hyperstages} displays the results obtained from comparing the models' performance across a range of stages, from one (SS-TCN) to five. The results revealed a consistent decrease of mean absolute difference values of \textit{all jumps}, as the number of stages increased up to four stages. Although the results of four-stage and five-stage showed similar results, the four-stage model was selected for fewer parameters.

\begin{table*}[!t]
\caption{Effects of the number of stages in MS-TCN}
\center
\label{tab:hyperstages}
\begin{tabular}{|l|lllllll|l|}
\hline
                                                            & \multicolumn{7}{c|}{Absolute mean difference}                                                                                                                                                                                                                 & \multicolumn{1}{c|}{\begin{tabular}[c]{@{}c@{}}correlation \\ coefficient (r)\end{tabular}} \\ \hline
                                                            & \multicolumn{1}{l|}{CMJ}          & \multicolumn{1}{l|}{block}        & \multicolumn{1}{l|}{smash}        & \multicolumn{1}{l|}{OS}           & \multicolumn{1}{l|}{squat} & \multicolumn{1}{l|}{dive} & \begin{tabular}[c]{@{}l@{}}All \\ jumps\end{tabular} & \begin{tabular}[c]{@{}l@{}}All \\ jumps\end{tabular}                                        \\ \hline
\begin{tabular}[c]{@{}l@{}}stage 1 \\ (SS-TCN)\end{tabular} & \multicolumn{1}{l|}{11.5}         & \multicolumn{1}{l|}{36.4}         & \multicolumn{1}{l|}{25}           & \multicolumn{1}{l|}{5.8}          & \multicolumn{1}{l|}{0.6}   & \multicolumn{1}{l|}{4.6}  & 35                                                   & 0.242                                                                                       \\ \hline
stage 2                                                     & \multicolumn{1}{l|}{\textbf{0.2}} & \multicolumn{1}{l|}{1.9}          & \multicolumn{1}{l|}{1.7}          & \multicolumn{1}{l|}{0.3}          & \multicolumn{1}{l|}{0.4}   & \multicolumn{1}{l|}{0.1}  & 1.6                                                  & 0.647                                                                                       \\ \hline
stage 3                                                     & \multicolumn{1}{l|}{0.3}          & \multicolumn{1}{l|}{\textbf{0.6}} & \multicolumn{1}{l|}{1.5}          & \multicolumn{1}{l|}{\textbf{0.1}} & \multicolumn{1}{l|}{0.5}   & \multicolumn{1}{l|}{0}    & 1.5                                                  & 0.72                                                                                        \\ \hline
stage 4                                                     & \multicolumn{1}{l|}{0.9}          & \multicolumn{1}{l|}{1.5}          & \multicolumn{1}{l|}{\textbf{0.2}} & \multicolumn{1}{l|}{0.2}          & \multicolumn{1}{l|}{0.8}   & \multicolumn{1}{l|}{0.4}  & \textbf{0.3}                                         & \textbf{0.884}                                                                                       \\ \hline
stage 5                                                     & \multicolumn{1}{l|}{1}            & \multicolumn{1}{l|}{1.8}          & \multicolumn{1}{l|}{0.4}          & \multicolumn{1}{l|}{0.2}          & \multicolumn{1}{l|}{0.6}   & \multicolumn{1}{l|}{0}    & 0.5                                                  & 0.882                                                                              \\ \hline
\end{tabular}
\end{table*}

Using the 4-stage architecture, a comparison was made between the MS-TCN and CNN-LSTM models, as presented in Table~\ref{tab:modelcomparison}. The MS-TCN outperformed CNN-LSTM on counting all jumps (r = 0.884 and 0.863, respectively). Also, it is important to note that CNN-LSTM required significantly more parameters to be trained effectively. Therefore, the MS-TCN model was ultimately selected for subsequent processing.\par

\begin{table*}[!t]
\caption{The absolute mean difference values of each activity, the correlation coefficients for all jumps, and the number of parameters for proposed MS-TCN and CNN-LSTM}
\center
\label{tab:modelcomparison}
\begin{tabular}{|l|lllllll|l|l|}
\hline
         & \multicolumn{7}{c|}{mean absolute difference of activity counts}                                                                                                                                                                       & \multirow{2}{*}{\begin{tabular}[c]{@{}l@{}}correlation \\ coefficient (r)\\ for all jumps\end{tabular}} & \multirow{2}{*}{\begin{tabular}[c]{@{}l@{}}number of \\ parameters\end{tabular}} \\ \cline{1-8}
         & \multicolumn{1}{l|}{CMJ} & \multicolumn{1}{l|}{block} & \multicolumn{1}{l|}{smash} & \multicolumn{1}{l|}{OS}  & \multicolumn{1}{l|}{squat} & \multicolumn{1}{l|}{dive} & \begin{tabular}[c]{@{}l@{}}all \\ jumps\end{tabular} &                                                                                                         &                                                                                  \\ \hline
MS-TCN   & \multicolumn{1}{l|}{0.6} & \multicolumn{1}{l|}{-0.6}  & \multicolumn{1}{l|}{0}     & \multicolumn{1}{l|}{1}   & \multicolumn{1}{l|}{0.9}   & \multicolumn{1}{l|}{0.1}  & 0.1                                                  & 0.884                                                                                                   & 118852                                                                           \\ \hline
CNN-LSTM & \multicolumn{1}{l|}{0.5} & \multicolumn{1}{l|}{1.2}   & \multicolumn{1}{l|}{0.1}   & \multicolumn{1}{l|}{1.2} & \multicolumn{1}{l|}{0.2}   & \multicolumn{1}{l|}{0.3}  & 0.3                                                  & 0.863                                                                                                   & 187975                                                                           \\ \hline
\end{tabular}
\end{table*}

\subsection{Overall jump count (comparing with VERT)}
\begin{figure}[!t]
\centerline{\includegraphics[width=0.9\columnwidth]{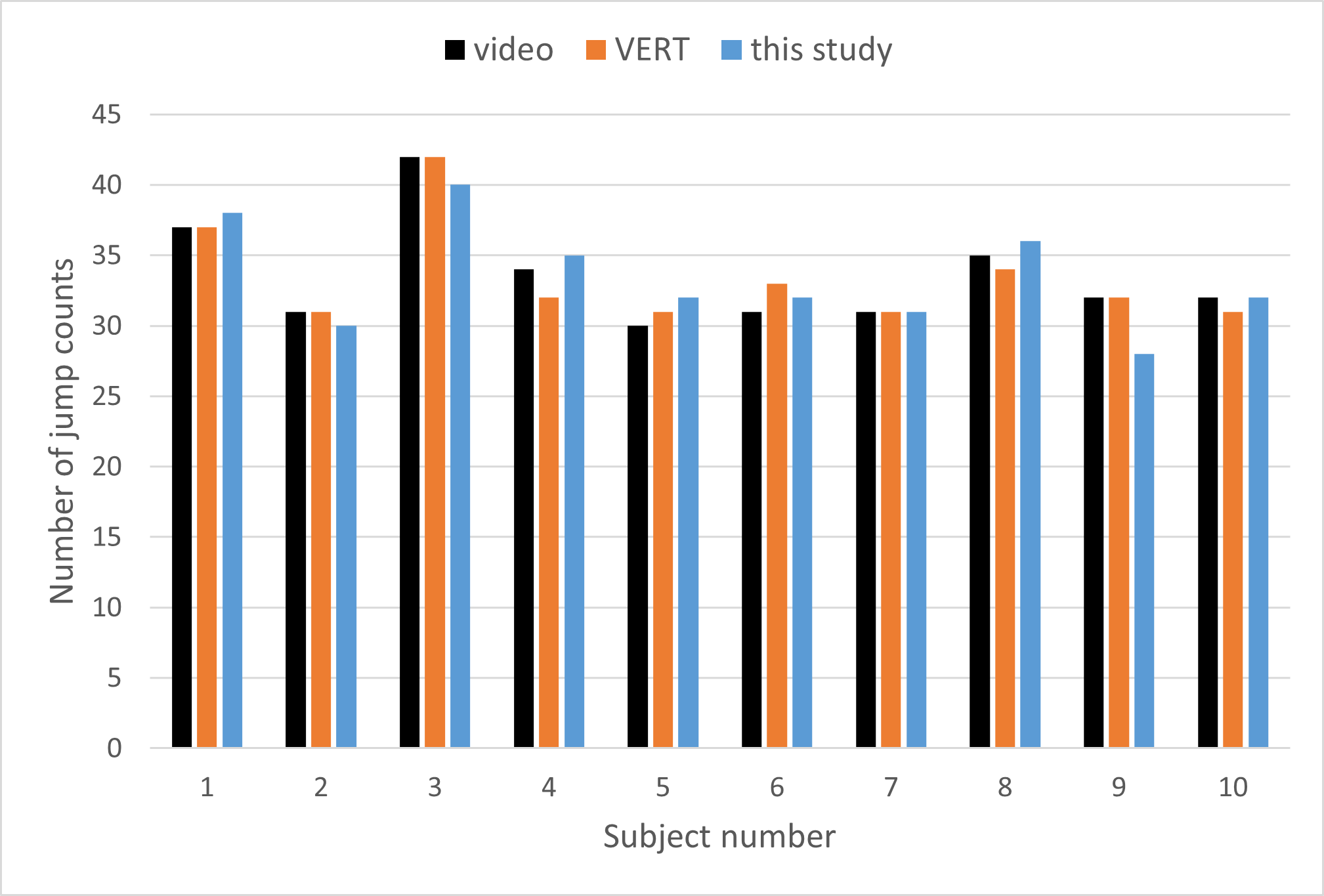}}
\caption{results of the classified number of jumps during the lab session using the proposed model, and the numbers given by the video and the VERT.}
\label{fig:VERT}
\end{figure}

During the lab session, VERT was also worn to count the jumps for comparison. Fig.\ref{fig:VERT} shows the jump counts of each subject by VERT and by this study. The figure shows that VERT obtained a smaller difference with video annotations for some subjects, compared with the proposed method. In general, the LOA by VERT was 0.1$\pm$2.08 ($r = 0.955$), which showed a smaller spread and higher correlation compared with the results of this study ($LOA = 0.1\pm3.40, r = 0.884$). 

\begin{figure}[!t]
\centerline{\includegraphics[width=.8\columnwidth]{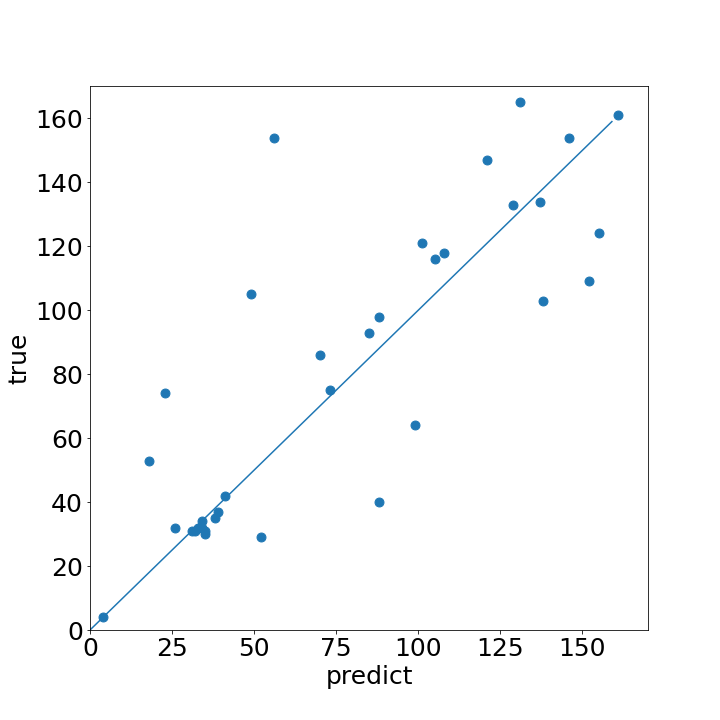}}
\caption{The number of predicted and true jump counts in the lab and training sessions, excluding \textit{OP/S} }
\label{fig:regression}
\end{figure}

Fig.~\ref{fig:regression} shows the true and predicted jump counts for each subject in both lab and training sessions (excluding \textit{OP/S}, as explained in section~\ref{subsec: results jump type}). It should be noted that there were several outliers where the outcomes were markedly inaccurate. Overall, the model underestimated the number of jumps.

\subsection{Jump types prediction}
\label{subsec: results jump type}

Using the selected MS-TCN model, the results of the predictions are presented in Table~\ref{tab:results}. The purpose of listing the results of the lab session was to compare them with those obtained during the training sessions.\par

\begin{table*}[!t]
\caption{Prediction results of lab sessions and training sessions}
\label{tab:results}
\center
\begin{tabular}{llllllllllll}
\hline
                        &         & CMJ                                                     & block                                                     & smash                                                     & OS                                                        & squat                                                     & dive                                                     & OP/S                                                       & hops                                                       & all jumps                                                & \begin{tabular}[c]{@{}l@{}}all jumps \\ (excl. OP/S)\end{tabular} \\ \hline \hline
\multicolumn{12}{c}{Lab session}                                                                                                                                                                                                                                                                                                                                                                                                                                                                                                                                                                                                                                    \\ \hline
\multirow{2}{*}{counts} & true    & 109                                                     & 112                                                       & 81                                                        & 33                                                        & 21                                                        & 43                                                       & \                                                          & \                                                          & 335                                                        & \                                                                   \\
                        & predict & 103                                                     & 118                                                       & 81                                                        & 43                                                        & 12                                                        & 44                                                       & \                                                          & \                                                          & 334                                                        & \                                                                   \\ \hline
\multicolumn{2}{l}{LOA}           & \begin{tabular}[c]{@{}l@{}}0.6$\pm$\\ 2.01\end{tabular} & \begin{tabular}[c]{@{}l@{}}-0.6$\pm$\\ 4.08\end{tabular}  & \begin{tabular}[c]{@{}l@{}}0$\pm$\\ 5.44\end{tabular}     & \begin{tabular}[c]{@{}l@{}}-1$\pm$\\ 4.00\end{tabular}    & \begin{tabular}[c]{@{}l@{}}0.9$\pm$\\ 2.74\end{tabular}   & \begin{tabular}[c]{@{}l@{}}-0.1$\pm$\\ 0.94\end{tabular} & \                                                          & \                                                          & \begin{tabular}[c]{@{}l@{}}0.1$\pm$\\ 3.40\end{tabular}    & \                                                                   \\ \hline \hline
\multicolumn{12}{c}{Training sessions}                                                                                                                                                                                                                                                                                                                                                                                                                                                                                                                                                                                                                              \\ \hline
\multirow{2}{*}{counts} & true    & \                                                       & 985                                                       & 1029                                                      & 612                                                       & 448                                                       & 111                                                      & 1797                                                       & 1635                                                       & 4423                                                       & 3074                                                                \\
                        & predict & \                                                       & 971                                                       & 833                                                       & 518                                                       & 419                                                       & 74                                                       & 570                                                        & 928                                                        & 3311                                                       & 2741                                                                \\ \hline
\multicolumn{2}{l}{LOA}           & \                                                       & \begin{tabular}[c]{@{}l@{}}0.54$\pm$\\ 17.67\end{tabular} & \begin{tabular}[c]{@{}l@{}}7.54$\pm$\\ 23.00\end{tabular} & \begin{tabular}[c]{@{}l@{}}3.62$\pm$\\ 17.62\end{tabular} & \begin{tabular}[c]{@{}l@{}}1.12$\pm$\\ 26.81\end{tabular} & \begin{tabular}[c]{@{}l@{}}1.42$\pm$\\ 4.11\end{tabular} & \begin{tabular}[c]{@{}l@{}}47.19$\pm$\\ 55.65\end{tabular} & \begin{tabular}[c]{@{}l@{}}26.15$\pm$\\ 36.43\end{tabular} & \begin{tabular}[c]{@{}l@{}}66.07$\pm$\\ 62.63\end{tabular} & \begin{tabular}[c]{@{}l@{}}17.85$\pm$\\ 36.70\end{tabular}          \\ \hline
\end{tabular}
\end{table*}

\begin{figure}[!t]
\centerline{\includegraphics[width=1\columnwidth]{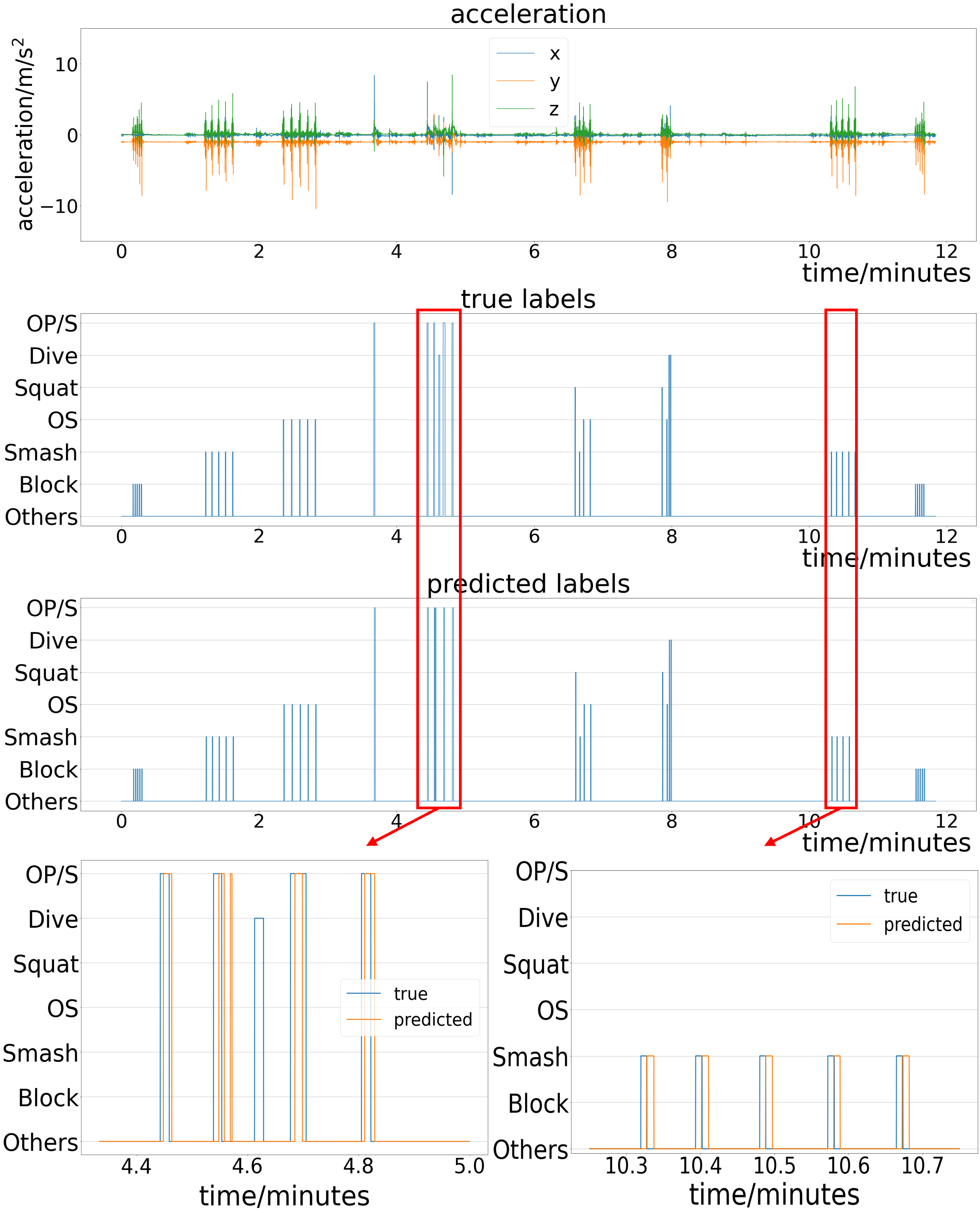}}
\caption{The acceleration signals, ground truth labels, and predicted labels of a subject in the lab session. The }
\label{fig:labresult}
\end{figure}

In the lab session, most jumps obtained high recognition performance. The mean difference values of \textit{CMJ}, \textit{block} and \textit{smash} were low. The true counts and predicted counts of \textit{OS} showed a significant difference (33 and 43, respectively). However, the prediction results for the overall jumps showed a low LOA value. Besides jumps, \textit{dive} could also be recognized with low errors. An example of the prediction results is shown in Fig.~\ref{fig:labresult}. The figure illustrates the fact that most jumps could be recognized, although different types of jumps could be mixed up. At the bottom of the figure, it also shows some examples of the detailed sample-wise prediction results, which illustrates the mismatches of the start/end time of each jump. The time shifts of such mismatches are variant from jump to jump. \par

Compared with the lab session, the performance decreased for the training sessions. Although the mean difference values of \textit{block}, \textit{smash}, \textit{OS}, \textit{squat} and \textit{dive} were low, the standard deviation was high. During the training sessions, the largest mean difference was associated with \textit{OP/S} (47.19), which had a significant impact on the results of \textit{all jumps}. In view of this, the last column of the table displays the results of \textit{all jumps} after excluding \textit{OP/S}.\par

\begin{figure*}[!t]
\centerline{\includegraphics[width=1.6\columnwidth]{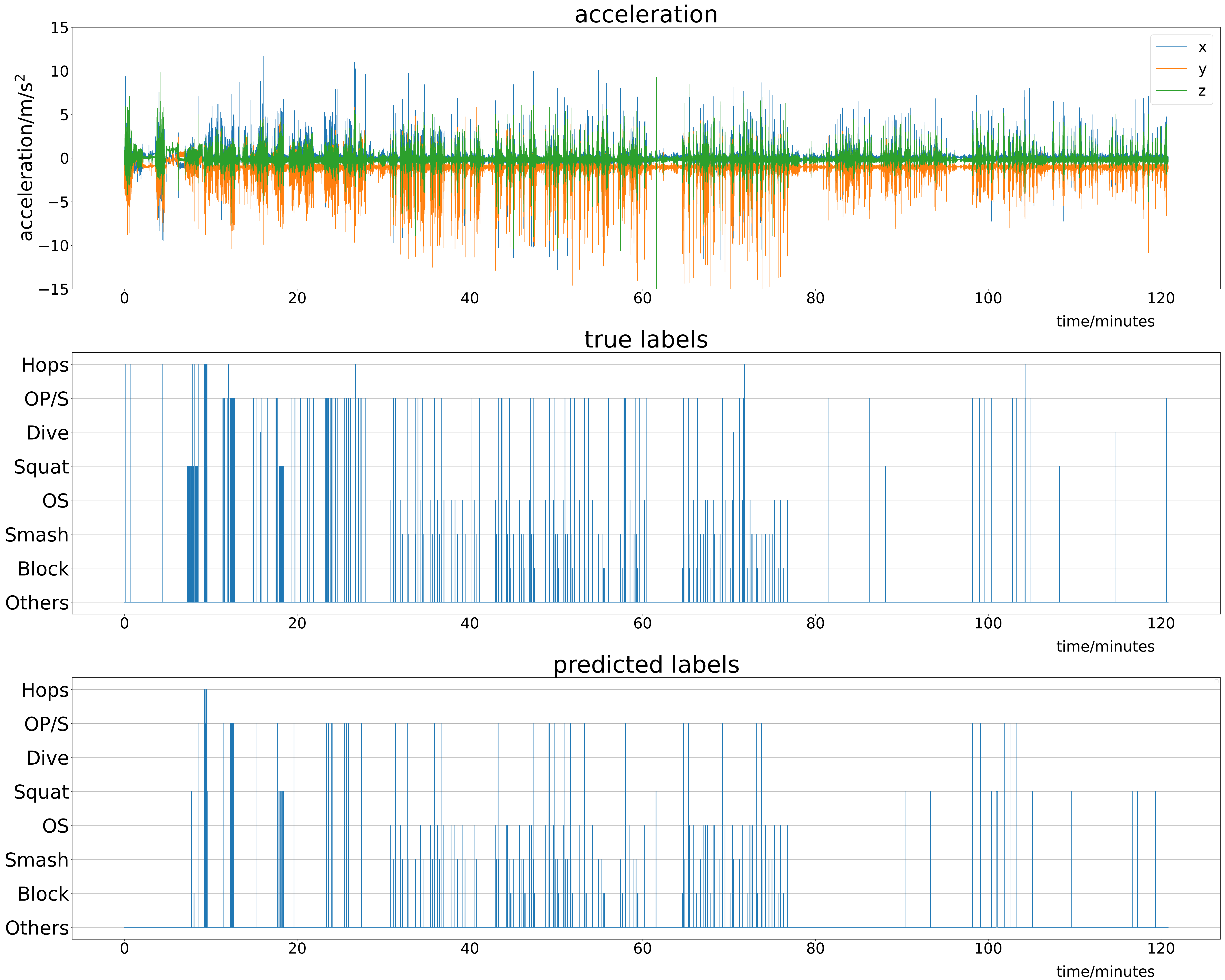}}
\caption{The acceleration signals, ground truth labels, and predicted labels of a subject in a training session}
\label{fig:trainingresult}
\end{figure*}

Fig.~\ref{fig:trainingresult} shows an example of the training sessions. Compared with Fig.~\ref{fig:labresult}, the signals in the training sessions were less clean because there were other intense activities involved in the sessions. As shown in Fig.~\ref{fig:trainingresult}, most false predictions happened in \textit{OP/S} and \textit{Hops}. Sometimes these activities were missed while sometimes they were classified as the other activities.\par

\section{Discussion}
\label{sec:discuss}

\subsection{The application of MS-TCN}

For MS-TCN, the number of stages influenced the refinement of the predicted time series. However, when the refinement performance reached the maximum, increasing the number of stages did not lead to better results anymore. This finding is consistent with the previous study comparing the number of stages in TCN \cite{farha2019ms}. \par

The 4-stage MS-TCN model obtained slightly better results to the CNN-LSTM model. Also, the CNN-LSTM model required more parameters to train, which increased computational costs. This finding is consistent with the previous study \cite{wang2022drinking}. In MS-TCN, the implementation of multiple dilated convolutional layers could increase the receptive field without increasing the filter size. An additional advantage of the MS-TCN model was that there were fewer hyperparameters to tune than CNN-LSTM, as shown in Table~\ref{tab:modelcomparison}. Therefore, it is easier to implement and train MS-TCN than CNN-LSTM.\par

Although the sample-wise prediction increased the resolution compared with the previous study using 0.5-s sliding window \cite{salim_towards_2020}, there were still mismatches of the start/end time of each jump (Fig.~\ref{fig:labresult}). One of the reasons might be the error caused by video synchronization. Therefore, some post-processing methods could be applied in the applications where an accurate start/end time is needed (e.g. when each jump needs to be analyzed for jumping force/height calculation \cite{damji2021using}). In the future, the estimation of jumping height/force prediction will be explored, based on the predicted jump occurrence in this study.

\subsection{Jumping prediction}

This study proved the possibility to recognize the volleyball jumping types using a single IMU. In the lab session, most activities obtained mean difference values smaller than one with low standard deviations. The poor results of \textit{OS} and \textit{squat} might be due to the small number of training samples of these types. The results of \textit{all jumps} showed slightly higher standard deviation and lower correlation compared with the results using VERT. However, the proposed system was able to recognize some jumping types using an IMU, which brings added value.\par

Using the same MS-TCN model in the training sessions, the performance was decreased compared with the lab session. The $M_{pd}$ values of \textit{block}, \textit{smash}, \textit{OS}, and \textit{squat} were still low but the $SD_{pd}$ were high. The LOA values (Table~\ref{tab:results}) and the results shown in Fig.~\ref{fig:regression} both illustrate the main problem of high variance among the subjects. The poor results could be caused by the fact that the small jumps were also involved in the dataset. Most previous studies excluded the jumps with height under 15cm for analysis \cite{benson_validation_2020, skazalski_valid_2018}. When the jumping height was too small, the IMU signals were not distinguishable from other dynamic activities such as running. In this study, the information of jumping height was unknown, which made the model always underestimate the number of jumps. The extremely poor results of \textit{OP/S} might also be reasoned by that most \textit{OP/S} jumps were small jumps.\par

The decreased prediction performance in the training sessions indicated two drawbacks of this study. First, the small jumps and big jumps were not annotated while collecting data. The small jumps are more challenging to be detected by a single IMU. Also, they are not as important as big jumps for musculoskeletal structures. Future studies should only focus on jumps beyond a certain height. Second, the system was not robust enough to generalize on different subjects, according to Fig.~\ref{fig:regression}. In the future, transfer learning could be applied to improve the training stage.

\section{Conclusion}
\label{sec:conclusion}

This paper proposed a multi-layer temporal convolutional network (MS-TCN) for sample-wise volleyball jumping classification using a single inertial measurement unit (IMU) mounted on the waist. The proposed MS-TCN model obtained better results than the state-of-the-art model with fewer hyperparameters for the classification of the volleyball jump. Although in the training sessions, the prediction results showed high variance among the subjects, the proposed system has the potential for monitoring volleyball players' jumps. The use of a single IMU sensor makes the system practical and cost-effective. Additionally, the proposed sample-wise prediction architecture provided a new approach to short-time activity recognition such as jumps.\par

In the future, a discrimination between jump height should be explored as well as the estimation of jumping height/force. Also, the estimation of jumping height/force prediction will be explored.

\section*{Acknowledgment}

This work was supported by the China Scholarship Council (CSC).

\bibliographystyle{ieeetr}
\bibliography{jsen}

\end{document}